\documentclass{article}

\usepackage{placeins}

\usepackage[utf8]{inputenc} % allow utf-8 input
\usepackage[T1]{fontenc}    % use 8-bit T1 fonts
\usepackage{hyperref}       % hyperlinks
\usepackage{url}            % simple URL typesetting
\usepackage{booktabs}       % professional-quality tables
\usepackage{amsfonts}       % blackboard math symbols
\usepackage{nicefrac}       % compact symbols for 1/2, etc.
\usepackage{amssymb}

\usepackage{amsmath}
\usepackage[ruled]{algorithm2e}

\usepackage{tikz}
\usetikzlibrary{arrows, automata}

\usepackage{amsthm}

\theoremstyle{definition}

\newtheorem{theorem}{Theorem}

\newtheorem{definition}{Definition}

\title{Explaining Black Boxes on Sequential Data using Weighted Automata}
\author{St\'ephane Ayache \\ \texttt{ firstName.lastName@lis-lab.fr}\\
QARMA team, LIS, Aix-Marseille University, France    
\and 
R{\'e}mi Eyraud \\ \texttt{firstName.lastName@lis-lab.fr}\\
QARMA team, LIS, Aix-Marseille University, France    
\and
No\'e Goudian \\ \texttt{firstName.lastName@lis-lab.fr}\\
QARMA team, LIS, Aix-Marseille University, France    
}
\begin{document}
\sloppy
\maketitle
%\maketitle

\begin{abstract}
Understanding how a learned black box works is of crucial interest for the future of Machine Learning. 
In this paper, we pioneer the question of the global interpretability of learned black box models that assign numerical values to symbolic sequential data. 
To tackle that task, we propose a spectral algorithm for the extraction of weighted automata (WA) from such black boxes. This algorithm does not require the access to a dataset or to the inner representation of the black box: the inferred model can be obtained solely by querying the black box, feeding it with inputs and analyzing its outputs.
Experiments using Recurrent Neural Networks (RNN) trained on a wide collection of 48 synthetic datasets and 2 real datasets show that the obtained approximation is of great quality.
\end{abstract}

\section{Introduction}
\label{sec:intr}
Recent successes of Machine Learning, in particular the so-called deep learning approach, and their growing impact on numerous fields, have risen questions about the induced decision process. Indeed, the most efficient models are often black boxes whose inner ruling system is not accessible to human understanding. However, explainability and interpretability are crucial  issues for the future developments of Machine Learning: to be able to explain how a learned black box works, or at least how it takes a decision on a particular datum, is a needed element for the development of the field~\cite{dosh17}, when it is not a legal requirement~\cite{rgpd}. 
 
A large debate on the meaning and limitations of explainability is currently occurring in Machine Learning~\cite{%keil14,
lipt16,dosh17}. We follow in this paper the recent survey from Guidotti et al.~ \cite{guid18} that describes two types of interpretability: the local one, that aims at explaining how a decision is taken on a given datum, and the global one, that tries to provide a general explanation of a black box model. In the framework of feed-forward models, several ideas have been studied. A well-known example of local interpretability would be to exhibit regions of a given image to justify its classification~\cite{cruz13,olah2018the%,DabkowskiGal2017RealB
} based on saliency detection methods and/or attention models. 
On the other hand, global interpretability could, for instance, take the form of the extraction of a decision tree or of a sparse linear model that maps black box inputs to its outputs~\cite{frei14,Ribeiro0G16}, usually based on distillation or compression approaches~\cite{hint15,
Bucilua06} that allow the learning of a possibly lighter and more interpretable student model from a teacher one.

In this paper, we focus on the harder problem of global interpretability for black boxes learned on sequential symbolic/categorial data. We also assume a general scenario where we do not have access to its training samples. 

%\paragraph{Related works.}
Few related works exist in that framework, most of which 
aim at extracting a deterministic finite state automaton (DFA) from a particular type of black boxes called recurrent neural network (RNN)~\cite{jaco05}.
% trained for binary classification~\cite{jaco05}.
For instance, \cite{gile92} propose a quantization algorithm to extract DFA from second order RNN; \cite{omli96} use a clustering algorithm on the output of a recurrent layer to infer a DFA; \cite{wang17} empirically compare different algorithms to get a DFA from a second order RNN; \cite{weis18} query a RNN to get pairs of (input, output) and use a recursive procedure to test the equivalence between their hypothesis and the RNN.

An important limitation of these works is that they all target RNN trained for binary classification, since DFA are non-probabilistic language models: as RNNs are not usually used for that task, this only gives some insights on the potential expressibility of RNNs, not on the interpretation of an existing RNN. 
An exception is the extension of the work of Omli \& Giles~ \cite{omli96} to the extraction of Weighted Automata (WA) from second-order RNN~\cite{leco12}. However, this last paper only focuses on a particular NLP task and lacks a general perspective.

The second important limitation of all these works is that they rely on finding a finite partition of the latent representation generated and used by the black box model: they all access the inside of the black box and differ mainly on the method to cluster this inner representation, from which they determine the states of the DFA they are extracting. %They also use this representation to get the transitions between the states.  

The work presented here handles a more common type of black boxes: we aim at extracting a finite state model from \textit{any} black box that computes a real valued function on sequential symbolic data. Moreover, our approach does not need to access the inside of the black box: we use it as an oracle, feeding it an input to get an output that is then analyzed. %Another important characteristic of our approach %that distinguishes the work presented here 
%is that the algorithm does not required the access to a learning dataset: all that is needed is the black box.
%In contrast to distillation methods, we don't consider having access to the black box training samples.

The core of the proposed algorithm relies on the use of a spectral approach that allows the extraction of a Weighted Automaton from any black box of the considered type. 
%defined on sequential categorial data. 
WAs~\cite{mohri2009} admit a graphical representation (see Figure~\ref{fig:PFA} for an example) while being more expressive than widely used formalisms, like Hidden Markov's Models% (HMM)
~\cite{deni08}. Furthermore, they have been heavily studied in theoretical computer science and thus both their behavior and their gist are well-understood~\cite{dros09}. 

After introducing the necessary preliminary definitions in Section~\ref{sec:prel}, we detail our algorithm for the spectral extraction of a WA from a black box in Section~\ref{sec:extr}. Section~\ref{sec:blac} presents the type of black boxes we use for the experiments, the RNNs, together with the used training protocol. Section~\ref{sec:expe} describes the experiments while Section~\ref{sec:res} details the obtained results. Finally, we discuss in Section~\ref{sec:disc} the limits and the potential impact of this work.

\section{Preliminaries}
\label{sec:prel}
\subsection{Elements of Language Theory}
\label{subsec:elem}
In theoretical computer science, a finite set of symbols is called an \textit{alphabet} and is usually denoted by the Greek letter $\Sigma$. Language theory mainly deals with finite sequences on an alphabet that are called \textit{strings} and are usually denoted by the letters $w$, $v$, or $u$. We denote the set of all possible strings on $\Sigma$ by $\Sigma^*$.
For instance, the alphabet can be the ASCII characters, the 4 main nucleobases of DNA, Part-of-Speech tags or lemmas from Natural Language Processing, or even a set of symbols obtained by the discretization of a time series~\cite{dimi10}.  

Throughout the paper, we will use other notions from language theory: the \textit{length} of a string $w$ is the number of symbols of the sequence (denoted $|w|$); the string of length zero is denoted $\lambda$; %we denote by $\Sigma^k$ the set of all the strings of length $k$ and by $\Sigma^{\leq k}$ the set of all the strings up to length $k$; 
given 2 strings $u$ and $v$ we note $uv$ their concatenation; if a string $w$ is the concatenation of %strings 
$u$ and $v$, $w=uv$, we say that $u$ is a \textit{prefix} of $w$ and that $v$ is a \textit{suffix} of $w$.  

\subsection{Functions on sequences}
\label{subsec:func}
In this paper, we consider functions that assign real values to strings: $f: \Sigma^* \to \mathbb{R}$. These functions are known under the name of \textit{rational series}~\cite{saka2009}. In particular, probability distributions over strings are such functions.
Each of these functions is associated with a specific object that had been proven to be extremely useful:

\begin{definition}[Hankel Matrix~\cite{balle14}] 
Let $f$ be a rational series over $\Sigma^*$. The Hankel matrix of $f$ is a bi-infinite
matrix $\mathcal{H} \in \mathbb{R}^{\Sigma^*\times\Sigma^*}$ whose entries are defined as $\mathcal{H}(u,v) = f(uv)$, $\forall u,v \in\Sigma^*$. Rows are thus indexed by prefixes and columns by suffixes.
\end{definition}

For obvious reasons, only finite sub-blocks of Hankel matrices are of interest. An 
easy way to define such sub-blocks is by using a \textit{basis} $\mathcal{B} = (\mathcal{P},\mathcal{S})$, where $\mathcal{P},\mathcal{S}\subseteq \Sigma^*$. 
If we note $p = |\mathcal{P}|$ and $s = |\mathcal{S}|$, 
the sub-block of $\mathcal{H}$ defined by $\mathcal{B}$ is the matrix $H_\mathcal{B}\in \mathbb{R}^{p\times s}$ with $H_\mathcal{B}(u, v) = \mathcal{H}(u, v)$ for any $u \in \mathcal{P}$ and $v \in \mathcal{S}$. 
We may write $H$ if the basis $\mathcal{B}$ is arbitrary or obvious from the context.

\subsection{Weighted Automata}
\label{subsec:wa}
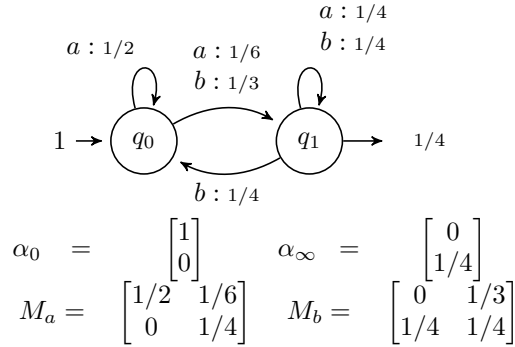
\begin{figure}[ht]
\begin{center}
\begin{tikzpicture}[->,>=stealth',shorten >=3pt,auto,node distance=2.2cm,semithick]
  \tikzstyle{every state}=[fill=none,draw=black,text= black]
  \tikzstyle{state with accepting}=[draw=none,fill=none]
  \node[initial left,state] (0)      [initial text={$1$}]              {$q_0$};
  \node[state]         (1) [right of=0] {$q_1$};

  \node[accepting] (T1) [right of=1]{};

  \path (0)edge                      [bend left]                           node {$\begin{matrix}a:{\scriptstyle 1/6}\\b:{\scriptstyle 1/3}\end{matrix}$} (1)
                 edge  [loop above,above left ] node {$a:{\scriptstyle 1/2}$} (0)
            (1) edge  [shorten >=31pt,right ] node {${\scriptstyle 1/4}$} (T1)
                 edge  [loop above ,above right] node {$\begin{matrix}a:{\scriptstyle 1/4} \\b:{\scriptstyle 1/4}\end{matrix}$} (1)
                  edge [bend left]  node {$b:{\scriptstyle 1/4}$} (0);
\end{tikzpicture}
   \hspace{1cm} 
$\begin{matrix}
\alpha_0\;\;\;=&\begin{bmatrix}1\\ 0 \end{bmatrix} & \alpha_\infty\;\;=&\begin{bmatrix}0\\1/4\end{bmatrix} \\
M_a=&\begin{bmatrix} 1/2&1/6 \\ 0&1/4 \end{bmatrix} &
M_b=&\begin{bmatrix} 0&1/3 \\ 1/4 &1/4 \end{bmatrix}\\
\ & \ & \ \\
\ & \ & \ \\
\ & \ & \ \\
\ & \ & \ \\
\ & \ & \ \\ 
\ & \ & \ \\ 
\ & \ & \ 
\end{matrix}$
\vspace{-2.5cm}
\end{center}
\caption{A WA in its graphical form and its equivalent linear representation.} 
\label{fig:PFA}
\end{figure}

%The following definitions are adapted from Mohri~\cite{mohri2009} and Balle~et~al.~\cite{balle14}:
\begin{definition}[Weighted automaton~\cite{mohri2009}]
A weighted automaton (WA) is a tuple $\langle \Sigma, Q, \mathcal{T}, \gamma, \rho \rangle$ such that:
%\begin{itemize}
%\item 
$\Sigma$ is a finite alphabet; 
%\item
$Q$ is a finite set of states; 
%\item 
%$I\subseteq Q$ is the set of initial states; 
%\item 
%$F\subseteq Q$ is the set of final states; 
%\item 
$\mathcal{T}: Q\times \Sigma \times Q \to \mathbb{R}$ is the transition function; 
%\item 
$\gamma: Q \to \mathbb{R}$ is an initial weight function; 
%\item 
$\rho: Q \to \mathbb{R}$ is a final weight function.
%\end{itemize}
\end{definition}
% A transition is usually denoted $(q_1,\sigma, p, q_2)$ instead of $\mathcal{T}(q_1,\sigma,q_2)=p$.
% We say that two transitions $t_1=(q_1,\sigma_1, p_1, q_2)$ and $t_2=(q_3,\sigma_2, p_2, q_4)$ are consecutive if $q_2=q_3$. A path $\pi$ is an element of $\mathcal{T}^*$ made of consecutive transitions. 
% We denote by $o[\pi]$ its origin and by $d[\pi]$ its destination. The weight of a path is defined by $\mu(\pi)=\gamma(o[\pi])\times\omega\times\rho(d[\pi])$ 
% where $\omega$ is the multiplication of the weights of the constitutive transitions of $\pi$.
% We say that a path $(q_0, \sigma_1, p_1, q_1)\ldots (q_{n-1}, \sigma_n, p_n, q_n)$ reads a string $w$ if $w=\sigma_1\ldots \sigma_n$.
% The weight of a string $w$, that is, the real value assigned by the WA to $w$, is the sum of the weights of the paths that read $w$.
A weighted automaton assigns weights to strings, that is, it computes a real value to each element of $\Sigma^*$.
WAs admit an equivalent representation using linear algebra:
\begin{definition}[Linear representation~\cite{%deni08,
balle14}]
A~\textit{linear representation} of a WA $A$ is a triplet $\langle \alpha_0, (M_\sigma)_{\sigma\in\Sigma}, \alpha_\infty \rangle$ where %the dimension of each element is the number of states of the automaton, 
the vector $\alpha_0$ provides the initial weights (\textit{i.e.} the values of the function $\gamma$ for each state), the vector $\alpha_\infty$ is the terminal weights (\textit{i.e.} the values of function $\rho$ for each state), and each matrix $M_\sigma$ corresponds to the $\sigma$-labeled transition weights ($M_\sigma(q_1,q_2)=p \Longleftrightarrow \mathcal{T}(q_1,\sigma,q_2)=p$). 
\end{definition}
Figure~\ref{fig:PFA} shows the same WA using the two representations. In what follows, we will confound the two notions and consider that WAs are defined in terms of linear representations.
 
% A series $r$ over an alphabet $\Sigma$ is a mapping $r: \Sigma^* \to \mathbb{R}$. A series $r$ over $\Sigma^*$ is \textit{rational} if there exists an integer $k \geq 1$, vectors $I,T \in \mathbb{R}^k$, and matrices
% $M_\sigma \in \mathbb{R}^{k\times k}$ for every $\sigma\in\Sigma$, such that for all $u = \sigma_1\sigma_2 \ldots \sigma_m \in \Sigma^*$,

To compute the weight that a WA $A$ assigns to a string $w=\sigma_1 \sigma_2\ldots\sigma_m$ using a linear representation, it suffices to compute the product $A(w)=\alpha_0^\top M_w \alpha_\infty = \alpha_0^\top M_{\sigma_1} M_{\sigma_2}\ldots M_{\sigma_m}\alpha_\infty$. This can be interpreted as a projection into $\mathbb{R}^r$, where $r$ is the number of states of $A$, followed by a inner product~\cite{rabu17}. Indeed, $\alpha^1=\alpha_0^\top$ is a vector that corresponds to the initial projection; each of the following steps computes a new vector $\alpha^{i+1}$ in the same space, moving from one vector to the next one by computing the product with the corresponding symbol matrix ($\alpha^{i+1}=\alpha^i M_{\sigma_i})$; the final projection is 
$\alpha^{m+1}=\alpha_0^\top M_w$; the output of of the WA on $w$ is given by the inner product $\langle
\alpha^{m+1}, \alpha_\infty\rangle$.

If a WA computes a probability distribution over $\Sigma^*$, it is called a stochastic WA. In this case, it can easily be used to compute the probability of each symbol to be the next one of a given prefix~\cite{balle14}: the probability of $\sigma$ being the next symbol of the prefix $w$ is given by %$\alpha^{i\top} M_\sigma \tilde\alpha_\infty$, where $\tilde\alpha_\infty = (I - (\sum_{\sigma\in\Sigma}M_\sigma))^{-1}\alpha_\infty$, and $\alpha^{i}$ is recursively computed as $\alpha^1 = \alpha_0$ and $\alpha^{i+1} = \alpha^{i\top} M_{\sigma_i}$.
$\alpha_0^\top M_w M_\sigma \tilde\alpha_\infty = \alpha^{|w|+1} M_\sigma \tilde\alpha_\infty$, where $\tilde\alpha_\infty = (Id - (\sum_{\sigma\in\Sigma}M_\sigma))^{-1}\alpha_\infty$. %, and $\alpha^{i}$ is recursively computed as $\alpha^1 = \alpha_0^\top$ and $\alpha^{i+1} = \alpha^{i} M_{\sigma_i}$.

The following theorem is at the core of the spectral learning of WA~\cite{hsu2009,bailly2009} and of our approach:

\begin{theorem}[\cite{carl71,flie74}]
\label{thm:finiterank}
A function $f : \Sigma^* \to \mathbb{R}$ can be defined by a WA iff the rank of its Hankel matrix is finite. In that case this rank is equal to the minimal number of states of any WA that computes $f$.
\end{theorem}

\section{Extracting WA from a black box}
\label{sec:extr}
%In this section, we describe our approach: from a already trained black box, we fill a finite part of a Hankel matrix, then use a spectral approach to obtain a WA from the matrix. The following subsections detail the different steps of our proposition.  
We recall we want to extract a WA from a given black box. Our setting is such that we only have access to an already learned model --- the black box --- but not to its training samples. %We will generate strings by sampling iteratively over next symbol prediction
%%%%%%%%%%%%%%%%%%%%%%%%%%%%%%%%%%%%%%%%%
\subsection{From Hankel to WA}
\label{subsec:H2WA}
The proof of Theorem~\ref{thm:finiterank} is constructive: it provides a way to generate a WA from its Hankel matrix $\mathcal{H}$. Moreover, the construction can be used on particular finite sub-blocks of this matrix: the ones defined by a complete and prefix-close basis. Formally, a basis $\mathcal{B} = (\mathcal{P},\mathcal{S})$ is \textit{prefix-close} iff for all $w\in \mathcal{P}$, all prefixes of $w$ are also elements of $\mathcal{P}$; $\mathcal{B}$ is \textit{complete} if the rank of the sub-block $H_\mathcal{B}$ is equal to the rank of $\mathcal{H}$.

Explicitly, from such sub-block $H_\mathcal{B}$ of $\mathcal{H}$ of rank $r$, one can compute a minimal WA %for the corresponding rational series $r$ 
using a rank factorization $PS=H_\mathcal{B}$, with $P\in\mathbb{R}^{p\times r}$, $S\in\mathbb{R}^{r\times s}$. If we denote $H_\sigma$ the sub-block defines over $\mathcal{B}$ such that $H_\sigma(u,v)=\mathcal{H}(u\sigma, v)$, and $h_{\mathcal{P},\lambda}$ the $p$-dimensional vector with coordinates $h_{\mathcal{P},\lambda}(u)=\mathcal{H}(u,\lambda)$, and $h_{\lambda,\mathcal{S}}$ the $s$-dimensional vector with coordinates $h_{\lambda,\mathcal{S}}(v)=\mathcal{H}(\lambda,v)$, then
the WA $A=\langle \alpha_0, (M_\sigma)_{\sigma\in\Sigma}, \alpha_\infty\rangle$, with:
\begin{center}
$\alpha_0^\top=h_{\lambda,\mathcal{S}}^\top S^+$, $\alpha_\infty=P^+ h_{\mathcal{P},\lambda}$, and, for all $\sigma\in \Sigma$, $M_\sigma=P^+ H_\sigma S^+$,
\end{center}
is a minimal WA\footnote{As usual, $N^+$ denotes the Moore-Penrose pseudo-inverse~\cite{moore1920} 
of a matrix $N$.} whose Hankel matrix is exactly the initial one~\cite{balle14}.

This procedure is the core of the theoretically founded spectral learning algorithm~\cite{bailly2009,hsu2009,balle14}, where the content of the sub-blocks is estimated by counting the occurrences of strings in a learning sample. 
Contrary to that, the work presented here uses 
%In this paper, we propose to use 
an already trained black box to compute $H_\mathcal{B}$ and $H_\sigma$ on a carefully selected basis $\mathcal{B}$.

\subsection{Proposed Algorithm}
\label{subsec:prot}

Our algorithm can be broken down into three steps: %from a black box model trained on some data%\footnote{The black boxes considered in this paper are Recurrent Neural Networks (RNN). We introduce them and describe their training in Section~\ref{sec:blac}.}
%,
first, we build a basis $\mathcal{B}$; 
second,
we fill the required sub-blocks $H_\mathcal{B}$ and $(H_\sigma)_{\sigma\in\Sigma}$  with the values computed by a black box model trained on some data; and third, we extract a WA from the Hankel matrix sub-blocks.% using a rank factorization. 

\begin{algorithm}
    \SetKwInOut{Input}{Input}
    \SetKwInOut{Output}{Output}
    %\DontPrintSemicolon
    %\underline{function Euclid} $(a,b)$\;
    \Input{Black box model $\mathcal{M}$, $p$, $s$ numbers of prefixes and suffixes, $r$ rank approximation}
    \Output{$A$, a Weighted Automaton}
    %\State{
        $(\mathcal{P},\mathcal{S})$ \hspace{27pt} $\leftarrow$ $Generate\_Basis(\mathcal{M}, p, s)$\; %}
     %\State{
        $H_\mathcal{B},(H_\sigma)_{\sigma\in\Sigma}$ $\leftarrow$ $Fill\_Hankels(\mathcal{M}, \mathcal{P},\mathcal{S})$\; %}
     %\State{
        $A$ \hspace{49pt}  $\leftarrow$ $Spectral\_Extraction(H_\mathcal{B},(H_\sigma)_{\sigma\in\Sigma}, r)$\tcp*{using equations from \ref{subsec:H2WA}}
      
      {
        \Return $A$\;
      }
    \caption{Extraction of a WA from a black box model on sequential data}
\end{algorithm}

Choosing the right basis $\mathcal{B}=(\mathcal{P}, \mathcal{S})$ 
%to define the right sub-block 
is an important task and different possibilities have been studied in the context of spectral learning~\cite{quat17,baillythesis}. For scalability reasons we chose to compute $Generate\_Basis()$ by sampling. 
If the black box is a generative device, we can use it to build a basis, for instance by recursively sampling a symbol from the next symbol distribution given by the black box.
Otherwise, we can obtain a basis by using the uniform distribution on symbols and a maximum length parameter\footnote{To present the more general results possible, this is the path followed in this paper.}, or by sampling a dataset if available.
Once a string is obtained, we add all its prefixes to $\mathcal{P}$ (to be prefix-close) and all its suffixes to $\mathcal{S}$. The process is reiterated until $|\mathcal{P}|\geq p$. If needed, the set of suffixes is then completed in the same way until $|\mathcal{S}|\geq s$.

%The first method that we use is to consider every possible string up to a given length $\ell$, so that $\mathcal{P} = \mathcal{S} = \Sigma^{\leq \ell}$ It gives satisfying results on settings with a small alphabet but this method can not be generalized to any size of alphabet, as the set of strings to consider becomes huge, forcing us to only consider a small maximal length for the strings, which does not carry enough informations. 
%Our second approach is intended to avoid this problem. Basing on \cite{baillythesis}, we know that randomly picking strings for our basis gives correct results. With parameters $p = |\mathcal{P}|$ and $s = |\mathcal{S}|$, $L_p$ and $L_s$ the set of lengths for prefixes and suffixes basis strings candidates respectively, we construct the prefix component $\mathcal{P}$ by randomly picking strings among $\bigcup\limits_{\ell\in L_p}\Sigma^\ell$, \textit{i.e.} all the possible strings with the relevant alphabet and of given sizes $L_p$. We take all the prefixes of these strings (doing so, we ensure that the base is prefix-closed) and we stop when the prefixes set has reached the desired size $p$. We use similar construction for suffix component $\mathcal{S}$. To consider all settings on an equal footing, only the second method is used. 

Once we have a basis $\mathcal{B}$, the procedure $Fill\_Hankels()$ uses the black box to compute the content of the sub-blocks: it queries each string made of a selected prefix and suffix to the black box and fill the corresponding cells in the sub-blocks with its answer.
% of the Hankel Matrix as presented in Section~\ref{subsec:H2WA}.

%Given the parameters $\mathcal{R}$ a trained RNN, $r$ the rank of the extracted WA, $p$ and $s$ the size of the prefix and suffix components of the base respectively, $L_p$ and $L_s$ the set of lengths for prefixes and suffixes basis strings candidates respectively, we extract a WA from $\mathcal{R}$:

%To compute the rank factorization and obtain a WA, we need a sub-block of the Hankel matrix of the function that we want to approximate, which is the one that $\mathcal{R}$ incarnates. To define a finite sub-block, we need to chose a prefix-closed basis ($\mathcal{P}$,$\mathcal{S}$). We construct its prefix component $\mathcal{P}$ by randomly picking strings among $\bigcup\limits_{l\in L_p}\Sigma^l$, \textit{i.e.} all the possible strings with the relevant alphabet and of given sizes $L_p$. We take all the prefixes of these strings (doing so, we ensure that the base is prefix-closed) and we stop when the prefixes set has reached the desired size $p$. We use similar construction for suffix component $\mathcal{S}$. Note that $p$ and $s$, are the most impacting parameters, both on the time consumption and the performance of the extraction. The impact of the $L_p$ and $L_s$ parameters is harder to understand, and dependent to the problem.

Finally, a rank factorization $H_\mathcal{B}=PS$ for a given rank parameter $r$, has to be obtained: the function $Spectral\_Extraction()$ performs a Singular Value Decomposition on $H_\mathcal{B}$ and truncated the result to obtain the needed rank factorization (see \cite{balle14} for details). It then generates a WA using the formulas described in Section~\ref{subsec:H2WA}.

\section{Black Box Learning}
\label{sec:blac}
This paper does not primarily focuses on learning a particular black box model: our approach is generic to any model that assigns real values to sequential symbolic data. However, to evaluate the quality of our algorithm for WA extraction, we need to have beforehand a learned model: we chose to use Recurrent Neural Network (RNN). %and this is the reason we introduce them here. 

\subsection{Recurrent Neural Network}
\label{subsec:RNN}
Recurrent Neural Networks (RNNs) are artificial neural networks designed to handle sequential data. To do so, a RNN incorporates an internal state that is used as memory to take into account the influence of previous elements of the sequence when computing the output for the current one.  

Two type of architecture units are mainly used: the widely studied Long Short Term Memory (LSTM)~\cite{hoch97} and the recent Gated Recurrent Unit (GRU)~\cite{cho14}. In both cases, these models realize a non-linear projection of the current input symbol into $\mathbb{R}^d$, where $d$ is the number of neurons on the penultimate layer: this vector is usually called the embedding or the latent representation of the part of the sequence seen so far. The last layer --- several layers can potentially be used here --- specializes the RNN to its targeted task from this final latent layer.

A RNN is often trained to perform the next symbol prediction task: given a prefix of a sequence, it outputs the probabilities for each symbol to be the next symbol of the sequence (a special symbol denoted $\rtimes$ [resp. $\ltimes$] is added to mark the start [resp. the end] of a sequence). %Outputting the final symbol means that the input sequence is not a prefix of a longer sequence but instead a complete sequence.
Notice that it is easy to use such RNNs to compute the probability given to a string $w=\sigma_1\sigma_2\ldots\sigma_m$: %one only needs to compute
$P(w) = P(\rtimes)P(\sigma_1|\rtimes)P(\sigma_2|\rtimes\sigma_1)\ldots P(\ltimes|\rtimes\sigma_1\sigma_2 \ldots\sigma_m)$

\subsection{Training}
\label{subsec:train}
We base the architecture on the work of \cite{shibataSpice}, who won the SPiCe competition~\cite{spice}, 
and of \cite{suts14}. The architecture is quite simple: it is composed of an initial embedding layer (with $3*|\Sigma|$ neurons), two GRU layers with tanh activation, two dense layers using ReLU activations, followed by a final dense layer with softmax activation composed of $|\Sigma|+1$ neurons.

Given this framework, we consider several hyper-parameters to tune. First, the number of neurons in the recurrent layers and the following dense layer: for GRU layers, we tested a number of neurons in ${30, 50, 120}$, the first following dense layer uses half of it, the second dense layer is set as the size of the input embedding layer. We trained our networks during 40 %300
epochs. We do witness expected over-fitting before this limit, confirming that this is an adequate number of maximum iterations. Finally, the model is trained using the categorical cross-entropy objective function.

For each problem, we keep the model (number of neurons, epoch's value) %%%%%%%%, size of the sliding window) 
that scores the best categorical cross-entropy on a validation set. The evaluation of this protocol shows good learning results (see top left plot of Figure~\ref{fig:global}) but it is also clear that better learning results could be obtained using RNNs on the chosen data. We decide to not push to its limits this learning part since it is not central in this work. Moreover, having RNNs with various learning abilities is interesting from the standpoint of the evaluation of the WA extraction: we expect the WA to be as good --- or, thus, as bad --- as the RNN.  

\section{Experimentations}
\label{sec:expe}

\subsection{Synthetic Data}
\label{subsec:data}
We chose to primarily evaluate our approach on the data from the PAutomaC learning competition~\cite{pautomac}. The goal of the competition was to learn a model from strings generated by a stochastic synthetic machine. The 48 instances of the competition equally featured Deterministic Probabilistic Finite Automata (DPFA), non-deterministic Probabilistic Finite Automata (PFA), and Hidden Markov's Models (HMM), which are all strictly less expressive than stochastic WA~\cite{deni08}. Their alphabet size range from 4 to 23 and their number of states from 6 to 73. They are intended, and the competition results proved it, to cover a wide range of difficulty levels. %Having access to the true target automaton gives an interesting point of comparison with the extracted automaton.
%Although they are purely synthetic data, we are convinced that they are relevant to evaluate the performance of our method, because they are not easy problems. Moreover, having access to the true target automaton gives an interesting point of comparison with the extracted automaton.

For each of the 48 problems, we have access to a training set (20\,000 or 100\,000 sequences), that we use to learn the RNN, a test set, and a description of the target machine.% that we mainly use to generate the validation set. 
%Of course the model description was only used during the evaluation, and not during RNN training nor extraction. Our use of the data ensured that no bias was induced by re-using the same validation sets.

% J'AI VIRE CA
%As already stated, we prepare the training set for the RNN by using a sliding windows of a given size. Two values are considered for this size: 10 and the length of the longest string in the training set. As a compromise between performance and running time, 60\,000 samples are randomly picked from the window-sliding extended training set.

\subsection{Real data}
\label{subsec:real}
In addition to synthetic data, we test our approach on two different real-world datasets: English verbs at character level from the Penn Treebank~\cite{penn} %, Spanish simplified POS sentences from the Ancora corpus~\cite{ancora}, and the sequences of the protein family PF00400 from the PFam database~\cite{pfam}. 
and discretized sensor signal of fuel consumption in trucks~\cite{verw11}. 
The first dataset contains 5\,987 learning examples on 33 different symbols and was problem 4 of the SPiCe competition~\cite{spice}. The second one is made of 20\,000 strings on an alphabet of size 18 and was used as PAutomaC Natural problem 2. In both cases, we use the preprocessed version of these data given by the corresponding competition. 

%For each of these tasks, we use the pre-processed data from the SPiCe competition: these datasets correspond to problem 4 , problem 8 (13\,903 examples, 48 symbols), and problem 10 (54\,932 examples, 20 symbols), respectively.

\subsection{Metrics}
\label{subsec:metr}
On the synthetic data, in order to not bias the evaluation by picking a particular data generator, we use two evaluation sets to compute the different metrics chosen to evaluate the quality of the WA extraction. $S_{test}$ is the test set given in PAutomaC, %; $S_{random}$ which is randomly generated using the target finite state machine with a \textit{high-temperature} modification~\cite{iba00} to its probabilities for each symbol in order to obtain less probable strings; 
and $S_{RNN}$ consists of generated sequences sampled using the learned RNN. $S_{test}$ contains 1\,000 strings %without repetition 
while $S_{RNN}$ is made of 2\,000 elements.

For the real data, we use a set $S_{test}$ of 2\,000 sequences that we randomly selected from the available data, and a set $S_{RNN}$ of 2\,000 sequences that we sampled using the RNN.

We evaluate the quality of %the approximation obtained by
the extraction using two types of metrics.

The first one consists in evaluating the similarity between the probability distribution $P_{RNN}$ of the RNN and the one of the WA, $P_{WA}$. To do so we compute the \textit{perplexity} (as in PAutomaC) and the \textit{Kullback-Leibler divergence} (KLD), where $\mbox{Perplexity}(P_{RNN}, P_{WA}) = 2^{-(\sum_{w \in EvalSet}P_{RNN}(w)log(P_{WA}(w)))}$, and $D_{KL}(P_{RNN}, P_{WA}) = \sum_{w \in EvalSet} P_{RNN}(w)log(\frac{P_{RNN}(w)}{P_{WA}(w)})$.

The second type of metrics aims at evaluating the proximity of the two models on the task that consists in guessing the next symbol in a sequence. We handle this part by computing the word error rate (WER)\footnote{KL and WER are only reported in the Appendix due to lack of place.}%~\cite{balle14}
, and normalized discounted cumulative gain (NDCG, as in SPiCe). This last metrics is given by: for each prefix $w^i$ of an element in an evaluation set, 
%$NDCG_n(\widehat{\sigma}_1^i,...,\widehat{\sigma}_n^i) = \frac{\sum_{k=0}^{n} \frac{P_{RNN}(\widehat{\sigma}_k^i|w^i)}{log(k+1)}}{\sum_{k=0}^{n} \frac{P_{RNN}(\sigma_k^i|w^i)}{log(k+1)}}$ 
$\mbox{NDCG}_n(\widehat{\sigma}_1^i,...,\widehat{\sigma}_n^i) = \sum_{k=0}^{n} \frac{P_{RNN}(\widehat{\sigma}_k^i|w^i)}{log(k+1)} / \sum_{k=0}^{n} \frac{P_{RNN}(\sigma_k^i|w^i)}{log(k+1)}$ 
where $\sigma^i_k$ [resp. $\widehat{\sigma}^i_k$] is the k-th most likely next symbol following $P_{RNN}$ [resp. $P_{WA}]$. The NDCG$_n$ score of an extraction is the sum of NDCG$_n$ on each prefix in the evaluation set, normalized by the number these prefixes. We compute NDCG$_1$ and NDCG$_5$ scores.

For completeness reasons, on the synthetic data we also compute these metrics to compare the RNN and the WA with the target machine, whose distribution over strings is denoted $P_{Target}$
%$P_{RNN}$ and $P_{WA}$ with the probability of the PAutomaC target machine $P_{Target}$. %Besides, we compute the $L_2$ automata distance~\cite{ball17} between the target machine and the extracted WA.
% Perplexity is computed on test set and random strings set between target and RNN, between RNN and extracted WA and between target and extracted WA. KLD is computed on test set and random strings set between target and RNN, between RNN and extracted WA and between target and extracted WA. It is also computed on random strings set between extracted WA and RNN. WER is computed on the test set for target, RNN and extracted WA and on RNN-generated strings for RNN and extracted WA. NDCG is computed with length 1 and 5 on test set between target and RNN, between RNN and extracted WA and between target and extracted WA. It is also computed with length 1 and 5 on the RNN-generated strings between RNN and extracted WA.
\subsection{Hyper-parameters for extraction}

We test different values for the hyper-parameters of the extraction algorithm: size $p$ and $s$ of the basis is taken between $(300,300)$, $(400,400)$, and $(800,800)$, %, $(1\,000, 1\,000)$, and $(1\,400, 1\,400)$, %$(1\,500, 1\,500)$, and $(2\,000, 2\,000)$,
the rank value ranges from 1 to 100. 

Some values of the size of the basis exceed what is reasonably computable on our limited computation capacities for some datasets. However, as it is shown in Section~\ref{sec:res}, small values already allow the extracted WA to be a great approximation of the RNN. 

All experiments are conducted using the Scikit-SpLearn toolbox~\cite{splearn} to handle WA and their extraction, and the Keras API~\cite{keras}, running on TensorFlow~\cite{tensorflow} backend, for RNN learning.

\section{Results}
\label{sec:res}
\subsection{Overall behavior}
\label{subseq:over}
\begin{figure}[!ht]
\begin{center}
%,height=160pt
\includegraphics[width=\textwidth]{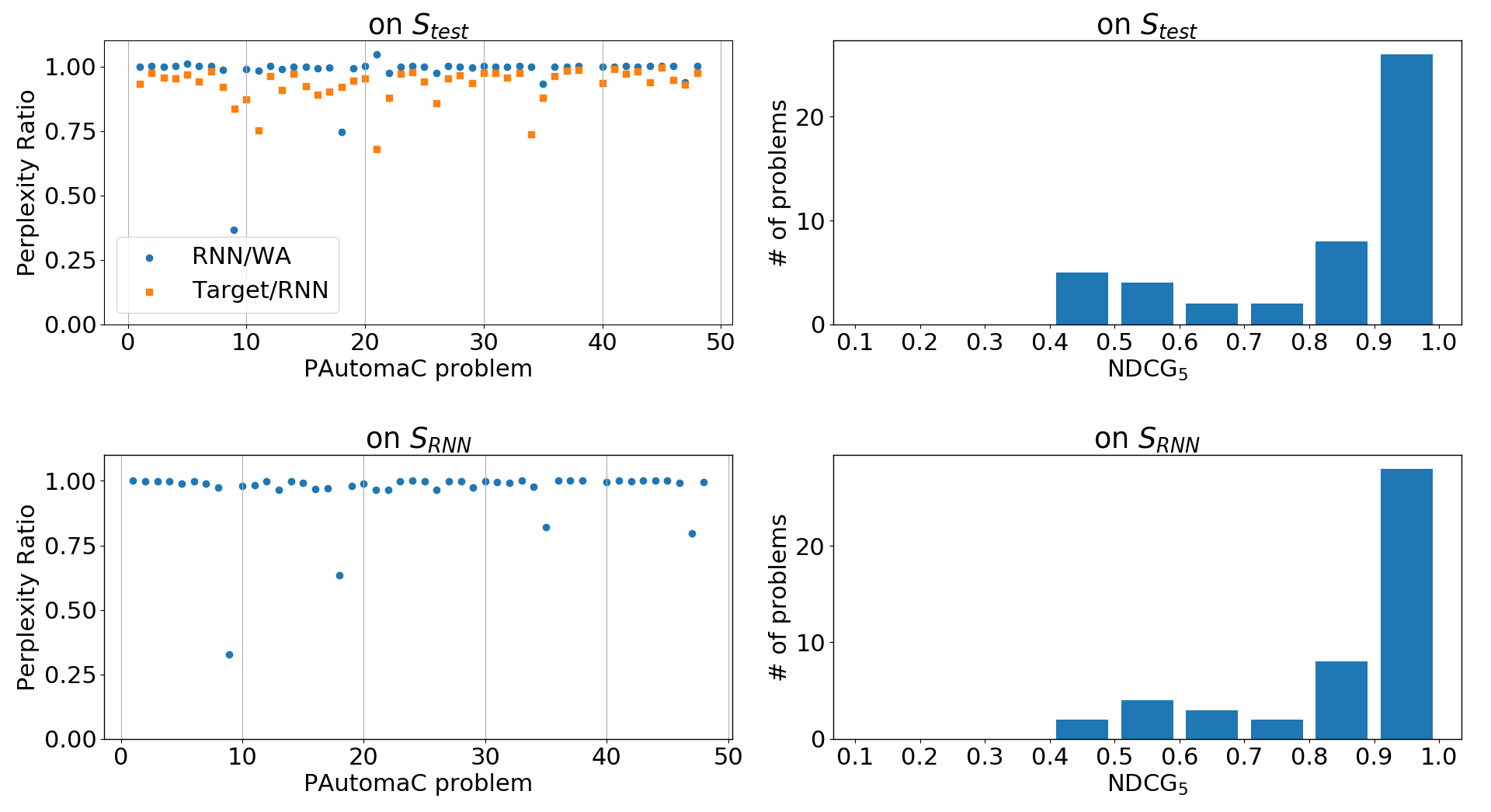}
\end{center}
\caption{Perplexity ratio and NDCG$_5$ for synthetic problems on the 2 evaluation sets. Top-left plot shows ratios
$\mbox{Perplexity}(P_{Target}, P_{RNN})/\mbox{Perplexity}(P_{Target}, P_{WA})$ in blue circles, and $\mbox{Perplexity}(P_{Target}, P_{Target})/\mbox{Perplexity}(P_{Target}, P_{RNN})$ in orange squares, 
both on the evaluation set $S_{test}$. 
%between the perplexity obtained by the target model with itself as reference and the perplexity obtained by the RNN with the target as reference (orange squares) and between the perplexity obtained by the RNN with the target as reference and the perplexity obtained by extracted WA with the target as reference (blue dots). 
Bottom-left plot shows ratio $\mbox{Perplexity}(P_{RNN}, P_{RNN})/\mbox{Perplexity}(P_{RNN}, P_{WA})$
on $S_{RNN}$.
The right plots show the number of problems per NDCG$_5$ decile on the 2 evaluation sets.} %between the perplexity obtained by the RNN with itself as reference and the perplexity obtained by the extracted WA with the RNN as reference (blue dots). Right side histograms shows the number of problems which obtain a certain range of NDCG$_5$ on $S_{test}$ and $S_{RNN}$ for top and bottom respectively.}
\label{fig:global}
\end{figure}

As the value of the best possible perplexity depends on the problem (for instance, the target perplexity for PAutomaC problem 47 is $4.12$ while for problem 2 it is $168.33$), we look at the ratio between the perplexity of the RNN and the one of the WA: 
$\mbox{Perplexity}(P_{Target}, P_{RNN})/\mbox{Perplexity}(P_{Target}, P_{WA})$. 
Figure~\ref{fig:global} shows the best obtained ratio and NDCG$_5$, both on $S_{RNN}$ and $S_{test}$\footnote{Generally the parameters that allow these different best scores are different for each of the four tasks. In Appendix, the same plots are given for each best parameters, showing the stability of these results.}. 

The perplexity ratio shows a remarkable proximity between RNNs and WAs on all but 2 datasets (problems 9 and 18). In addition, this closeness between the 2 distributions does not seem to depend on the quality of the learning process, given by the perplexity ratio between the PAutomaC target model and the RNN. In particular, we can notice that problems 11, 21 and 34 have a low Target/RNN perplexity ratio, whereas RNN/WA is closed to the optimum, meaning that WA extraction works also well for poor RNN performances. %\textcolor{blue}{We also remark that only one problem (number 18) conducted to a (much) poorer perplexity on the WA than RNN.}

%We witness a more important variability of the NDCG$_5$ score. This is likely to come from the inherent volatility of the metric and the difficulty to obtain a high score: for instance, the SPiCe competition use it and the best submissions on some problems hardly get $0.2$. We note that a large majority of the WA score higher than $0.9$ and only two less than $0.5$. This tends to indicate that the extracted WA is close to the RNN.  

Regarding the distribution of NDCG$_5$ scores over 48 PAutomaC problems, we note that a large majority of the WA score higher than $0.9$ and about $70\%$ of problems with a score higher than $0.8$. This %tends to indicate 
indicates that the WA estimations of next symbol probability are close to the RNN outputs.

%\textcolor{red}{A word on Target/RNN}
%\textcolor{blue}{Perplexity ratio and MDCG ratio (+ a word on "different best parameters" with ratio graph for best parameters of the other metric in appendix).}

%\textcolor{blue}{Conclusion for global: works well! (good approximation)}
\subsection{Influence of the hyper-parameters}
We analyze in this section the impact of the sizes of $\mathcal{P}$ and $\mathcal{S}$, as well as the rank for WA extraction on 2 synthetic and the 2 real datasets. 

\label{subec:}
\begin{figure}[ht]
\centering
%width=0.8\textwidth,height=90pt
\includegraphics[width=\textwidth,height=130pt]{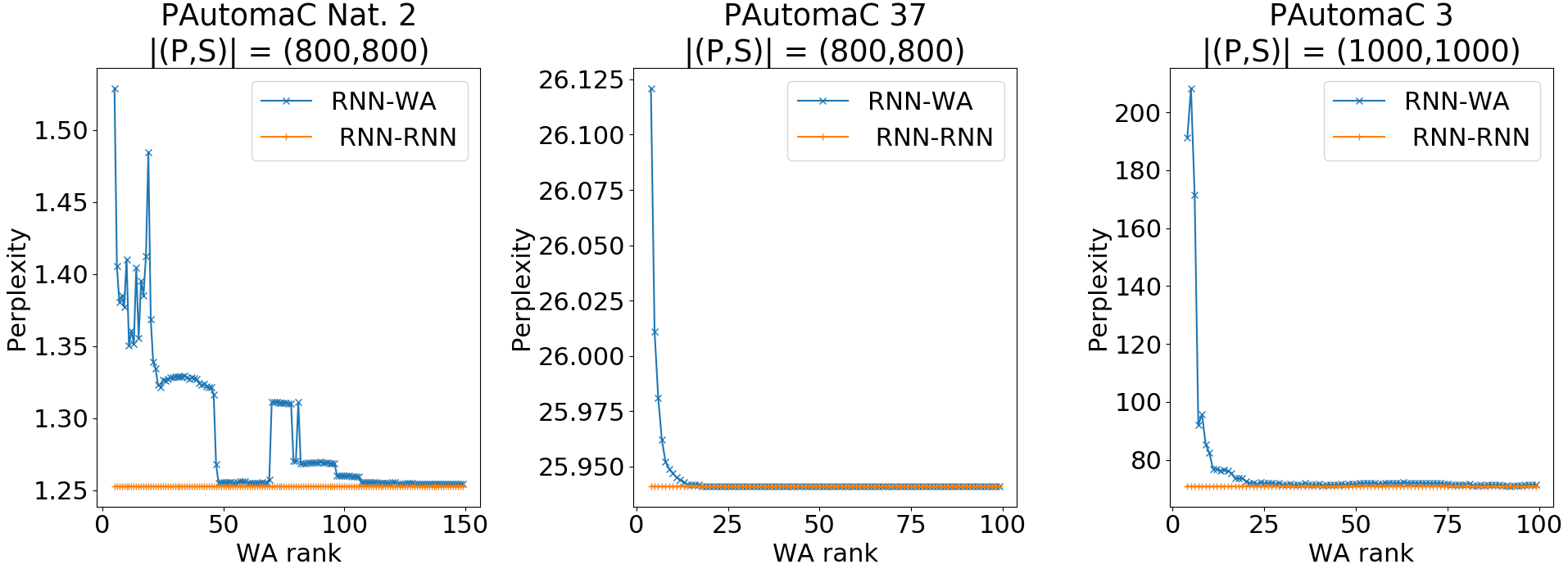}
\caption{Influence of the rank parameter on the perplexity of 3 problems. PAutomaC Nat. 2 is the second real dataset presented in Section~\ref{subsec:real}. PAutomaC 37 and PAutomaC 3 both correspond to a PFA target: the first one has 69 states on 8 symbols, the second 25 on 4.}
\label{fig:rank}
\end{figure}

Figure~\ref{fig:rank} shows the influence of rank on the quality of extracted WA, measured by the perplexity. We give both $\mbox{Perplexity}(P_{RNN}, P_{RNN})$ and $\mbox{Perplexity}(P_{RNN}, P_{WA})$. We notice that $\mbox{Perplexity}(P_{RNN}, P_{RNN})$ corresponds to the RNN's entropy, that is, the best possible score. %, where $\mbox{Perplexity}(P_{RNN}, P_{WA})$ could tend, with respect to a given problem. 
We do not show results with rank values less than 5 in order to make the plots readable: for instance, the perplexity at rank 1 was higher than 20, for PAutomaC Nat. 2.

As expected, it appears that the higher the rank is, the better the quality of WA extraction becomes. Notice however that reasonable perplexity is obtained for small rank values. Our extracted WAs seem almost optimal with as few as 25 states.

%a small rank can approximate the performance of RNN with a reasonable perplexity difference, as seen in the two plots on the right of figure~\ref{fig:rank}. For those two synthetic problems (PAutomaC 37 and PAutomac 3), the original automata used to generate data samples are both PFA, made of 69 states and 8 symbols (resp. 25 and 4). Our extracted WAs seem almost optimal with about 25 states.

On real problem PAutomaC Nat. 2, the quality of the extracted WA behaves more chaotically with rank variations, but still tends to converge to the optimum. We notice that variations are negligible since a perplexity of $1.35$ is already acceptable when the best possible is $1.25$.

% plus on a un gros rank, mieux c'est
% si on a un petit rank sur certins problemes et on est deja optimal
%\vspace{-0.2cm}
\begin{figure}[ht]
\centering
%,height=110pt
\includegraphics[width=\textwidth,height=110pt]{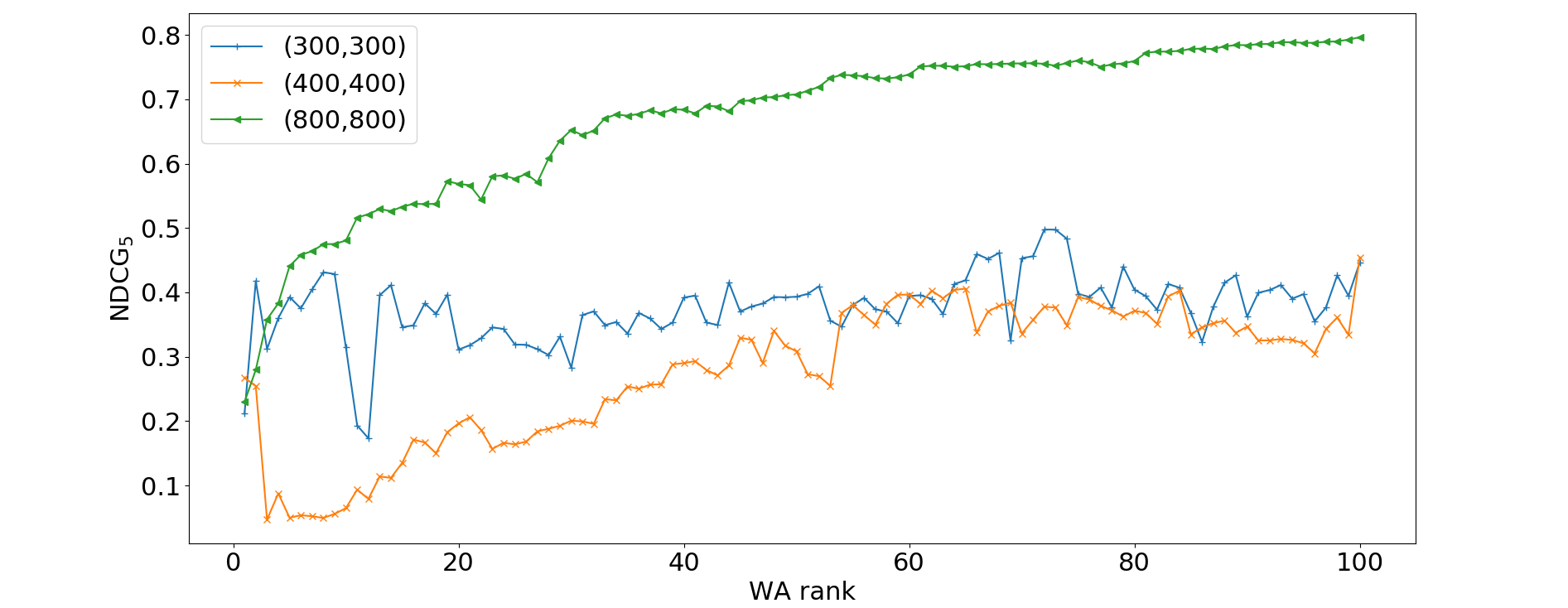}
\caption{Influence of the size of the basis and the rank on NDCG$_5$ for problem SPiCe 4 (NLP). Each curve correspond to a different size of basis.}
\label{fig:basis-size}
\end{figure}

Figure~\ref{fig:basis-size} illustrates the impact of the size of the basis $\mathcal{B}$ and of the rank used to extract the WA. It appears that the quality of the WA approximation, measured in terms of NDCG$_5$, seems to increase with the number of prefixes and suffixes in the basis. 
%using as many as possible prefixes and suffixes. 
The difference between configurations (300,300) and (400,400) is not significant, but doubling the basis size significantly improves the NDCG$_5$ score. This suggest that using even a larger basis might be useful on some tasks.

\section{Discussion}
\label{sec:disc}
Experiments show that our approach allows good approximations of black boxes, demonstrating that the linear projection defined by a WA can be close to the non-linear one of a RNN.
Furthermore, we want to emphasize the fact that we dis not chose the most favorable framework: for instance, using RNNs to generate the basis could lead to better suited basis and thus to better approximation (for this future work, we may need RNNs specifically trained to generate sequences~\cite{grav13} to avoid the autoregressive behavior). 
Using larger bases could also allow better results as our experiments tend to show (see Figure~\ref{fig:basis-size}): this was not doable in reasonable time using our configuration (CPU, 4 cores, \textasciitilde{}25Go RAM) but it is likely to not be a problem on state-of-the-art computers.

It is also worth noticing that, though we tested it on probabilistic models, the algorithm works on any black boxes that assign real values to sequences (or that can emulate this scheme, like the RNNs of our experiments) since WAs are not limited to probability distributions. %Besides, the analyzed black box does not need to be a generative model: the experiments presented here use the RNN to generate the basis, but preliminary results on uniformly sampled bases show acceptable approximations.

Another point that needs to be discussed is the interpretability of WAs: though they admit a graphical representation and are widely used in many fields, their non-deterministic nature can make them hard to read when the number of states increases. A first answer to that remark is that the algorithm described in this paper depends on a parameter, the rank, that can be tweaked: as the approximations for small rank values are already of acceptable quality, one can prefer readability to performance and chose a small rank value to obtain a small WA (see the example of a low rank extracted WA in Appendix%, figure~\ref{fig:extracted}
). Our algorithm can then be seen as a way to compute a limited development of a black box function into WAs: by fixing the rank, one decides how detailed, and thus how close to the learned model, the extracted WA has to be.   

A second answer to that point is that computing a weight (or the probability of the next symbol) is less expensive with WAs than with black boxes like RNNs. Indeed, the computation requires only matrix products, one per symbol in the input sequence, while non-linear models necessitate for each symbol several such products and the computation of non-linear functions. This is why the proposed algorithm 
share characteristics with 
%can be seen as a 
distillation processes~\cite{hint15}: from a complex, computationally costly model, it generates a simpler and more efficient one whose abilities are comparable.

%\textcolor{green}{Future works ?}

%\textcolor{blue}{transform the WA into a tree structure with backward transition (François?)}

%\textcolor{blue}{The algorithm works for black boxes that do not compute probability distribution (all that is needed: function from sequences to a real-value - or that can be transformed into this framework as the RNN presented here).}

%\textcolor{blue}{future work? limitations?}

%\textcolor{blue}{In certain cases, we observe that the extracted WA is doing better than the learned RNN on the initial learning problem (see the perplexity ratio of PAutomaC problem 21 for instance). This means that the extraction may correspond to a smoothing phase. Algorithms using both the WA and the RNN to infer a better model are thus worth to be studied.}
%\textcolor{blue}{readability of large WA vs limited development (already correct at rank 4)}

%\textcolor{blue}{2 different projections: non-linear embedding of RNN vs linear one of WA.}

%\textcolor{green}{Faster, low consumption, distillation}

\paragraph*{Acknowledgements.}We want to thanks Fran{\c c}ois Denis and Guillaume Rabusseau for fruitful discussions on WA and Benoit Favre for his help on RNN.
\small
\bibliography{RNN2WA}
\newpage
\normalsize
\section*{Appendix}
\subsection*{Example of an extracted WA}
\begin{figure}[ht]
\begin{center}
\includegraphics[scale=0.5]{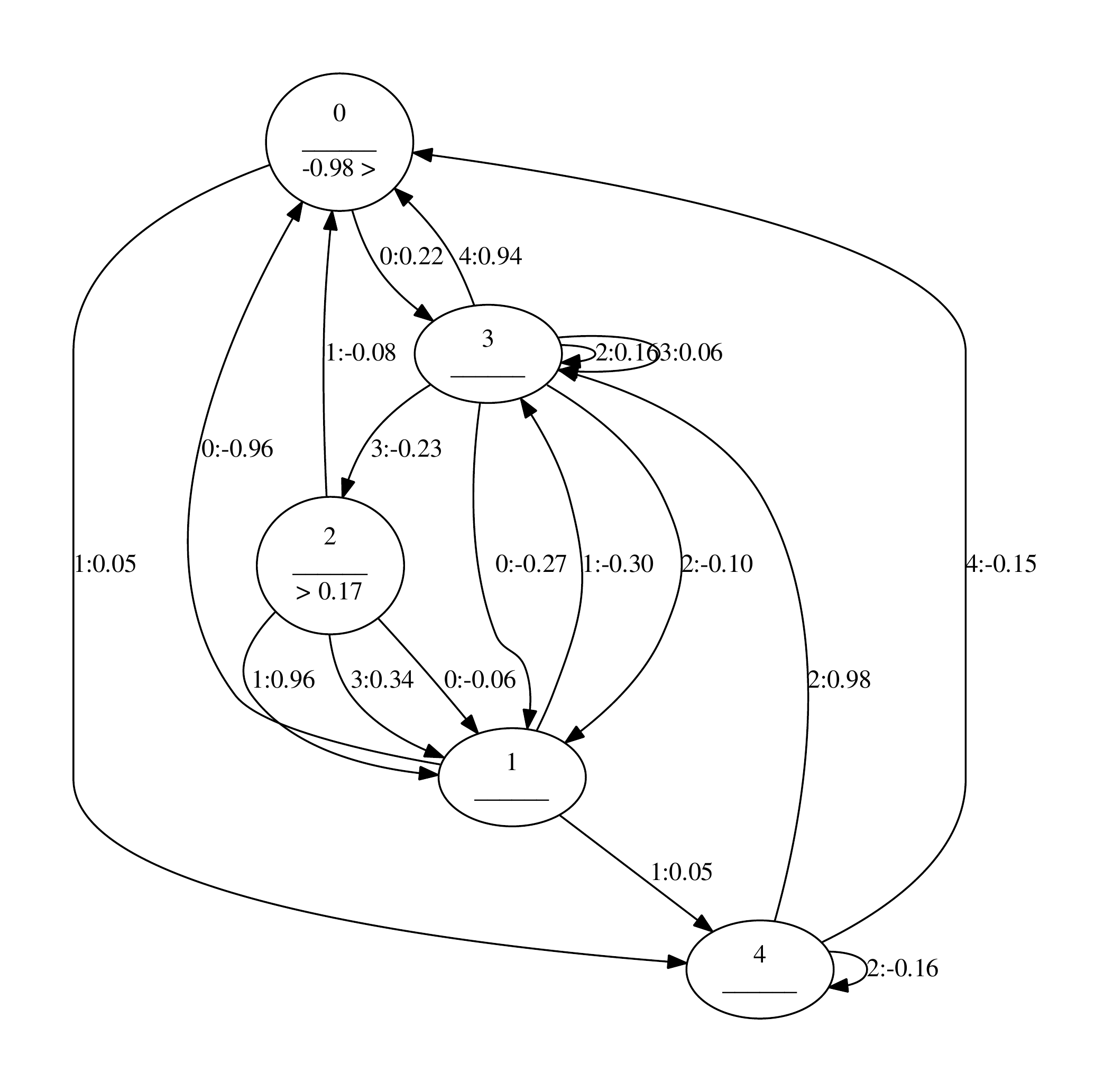}
\end{center}
\caption{WA extracted for problem 24, at rank 5, with basis size 800x800. Its perplexity ratio is $0.99849$ while its NDCG$_5$ is 0.99848. Input weights are given on state before the $>$ symbol, while output ones appear after $>$. Transitions with absolute weight under 0.05 are not shown.}
\label{fig:extracted}
\end{figure}

Figure~\ref{fig:extracted} gives the graphical representation on a WA extracted from a RNN trained on PAutomaC problem 24. This is not the best obtained WA on that dataset, but the metrics show that it is still a good approximation of the RNN. 

\subsection*{Metrics for best parameters}
\begin{figure}[ht] 
\begin{center}
\includegraphics[width=\textwidth]{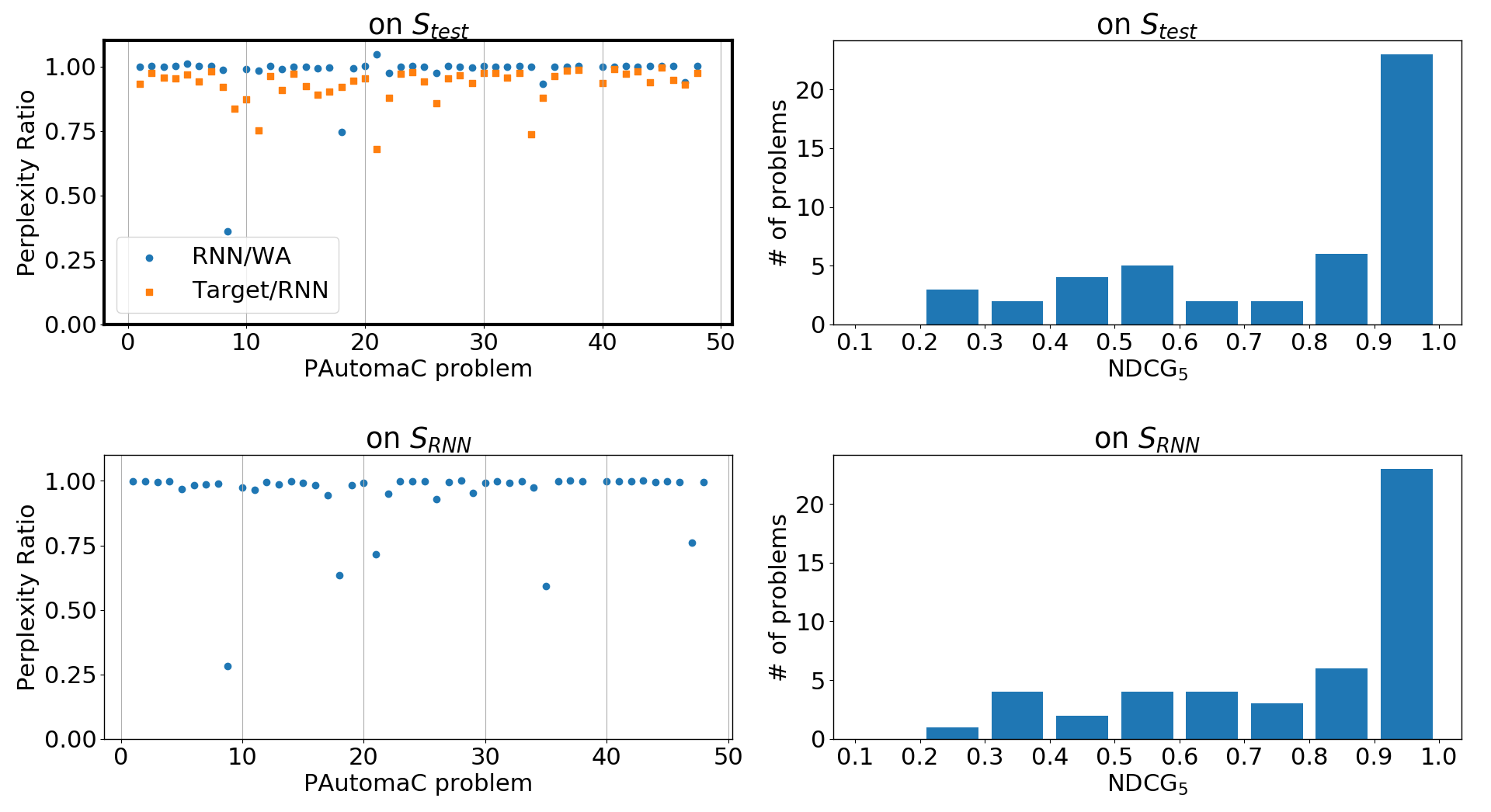}
\end{center}
\caption{Perplexity ratio and NDCG$_5$ for the PAutomaC problems on the 2 evaluation sets using the parameter scoring the best perplexity ratio on $S_{test}$.}
\label{fig:metrics1}
\end{figure}

\begin{figure}[h!]
\begin{center}
\includegraphics[width=\textwidth]{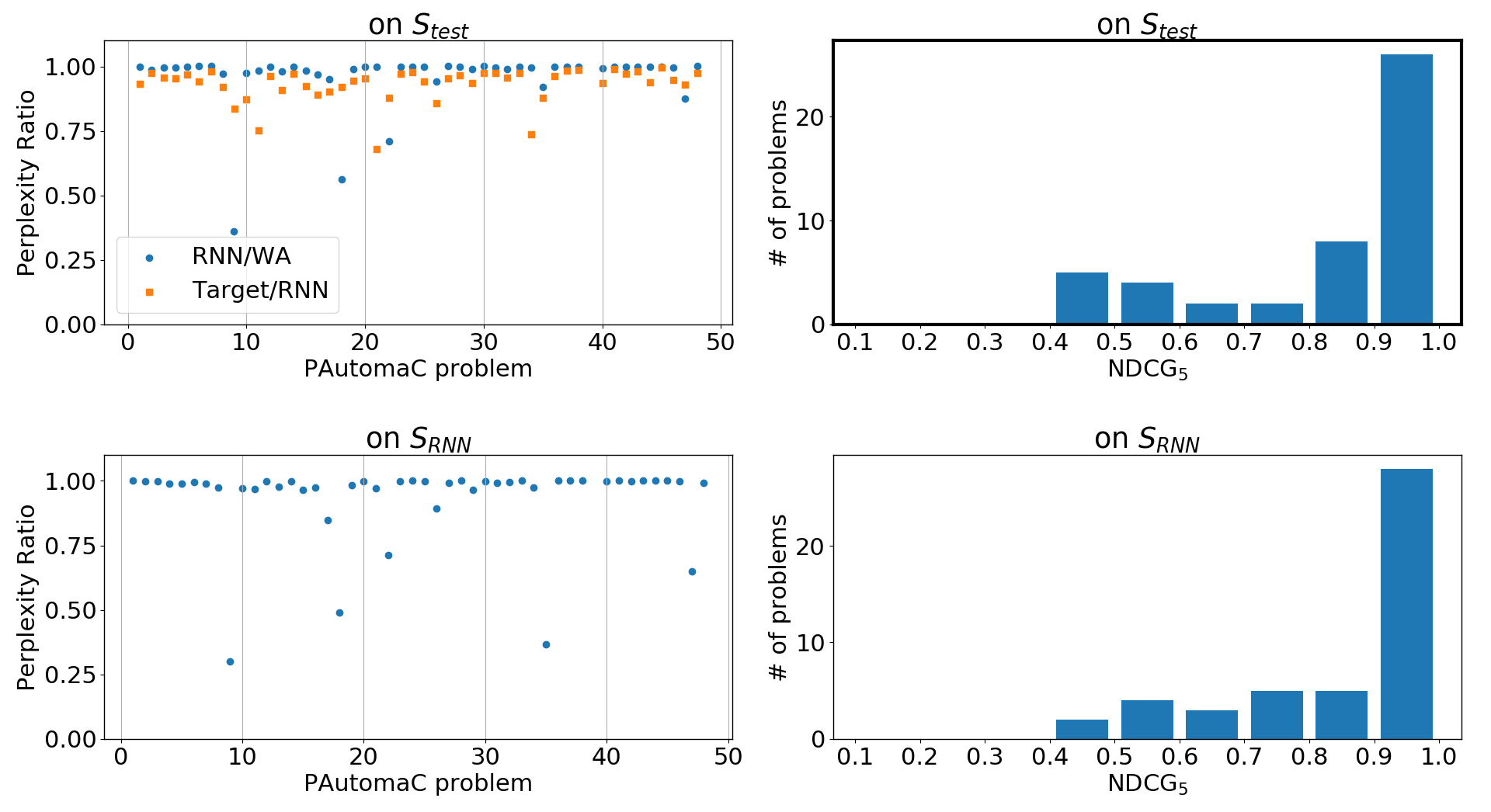}
\end{center}
\caption{Perplexity ratio and NDCG$_5$ for the PAutomaC problems on the 2 evaluation sets using the parameter scoring the best NDCG on $S_{test}$.}
\label{fig:metrics2}
\end{figure}

\begin{figure}[!htb]
\begin{center}
\includegraphics[width=\textwidth]{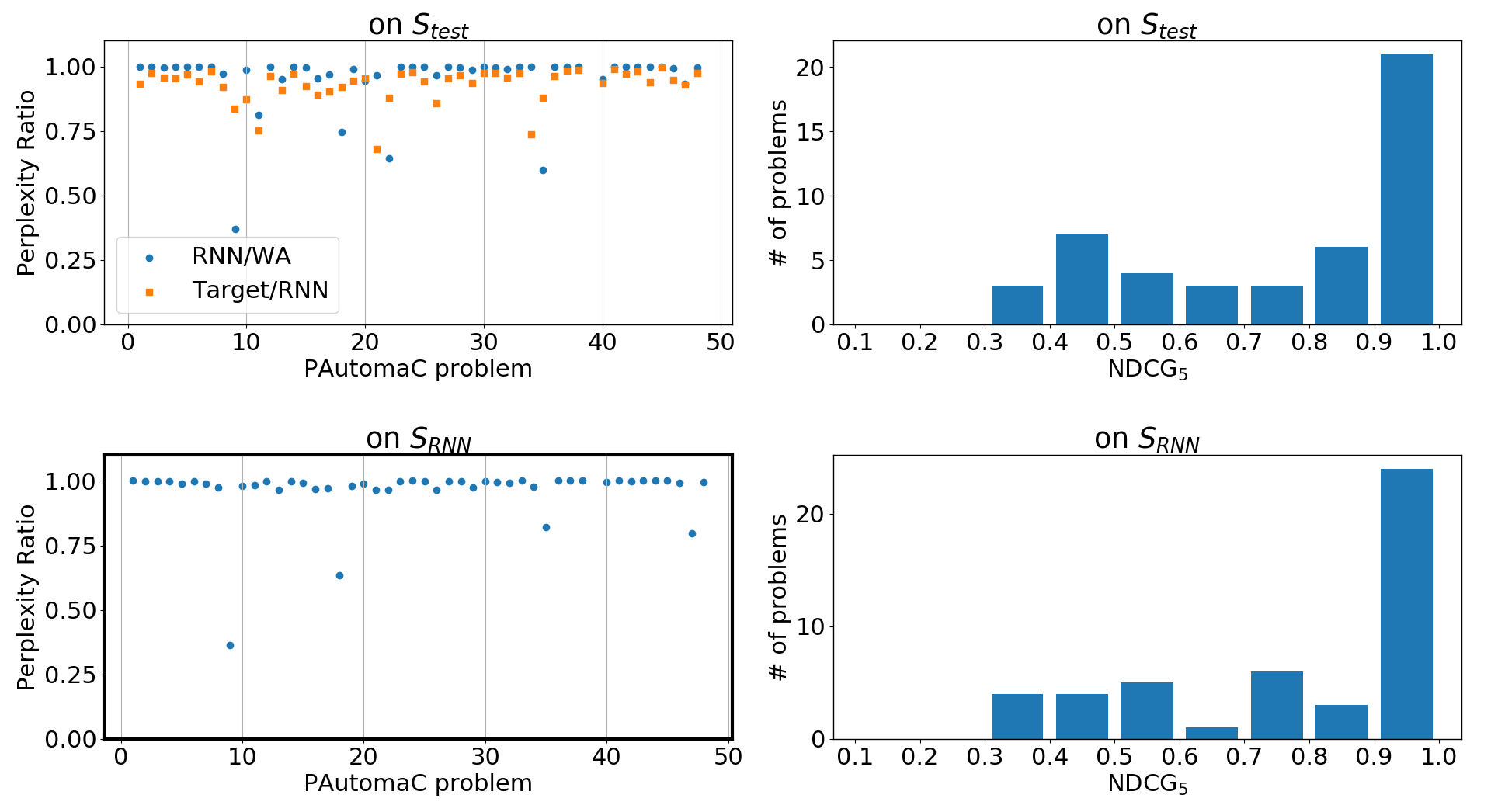}
\end{center}
\caption{Perplexity ratio and NDCG$_5$ for the PAutomaC problems on the 2 evaluation sets using the parameter scoring the best perplexity ratio on $S_{RNN}$.}
\label{fig:metrics3}
\end{figure}

\begin{figure}[ht]
\begin{center}
\includegraphics[width=\textwidth]{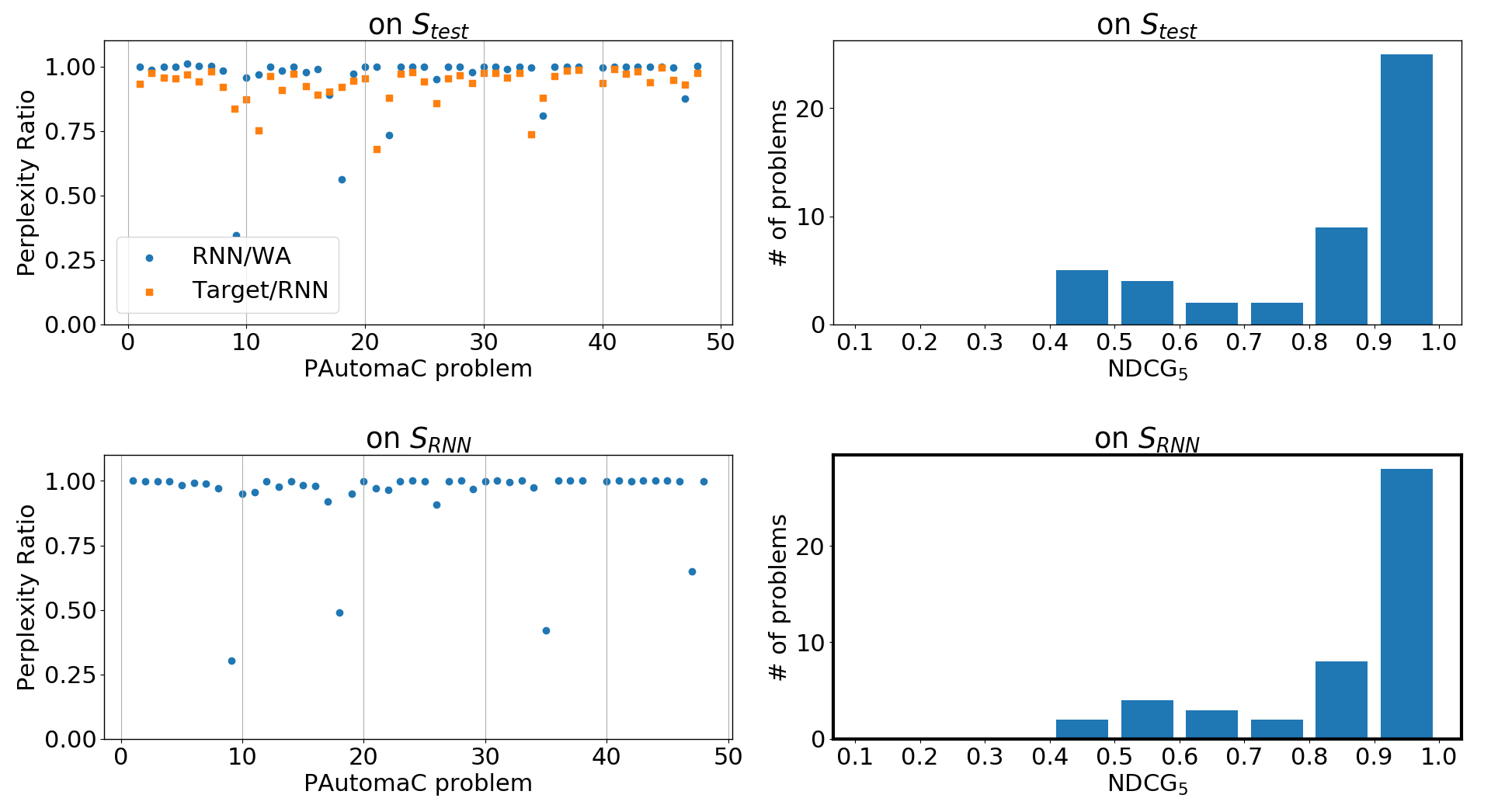}
\end{center}
\caption{Perplexity ratio and NDCG$_5$ for the PAutomaC problems on the 2 evaluation sets using the parameter scoring the best NDCG on $S_{RNN}$.}
\label{fig:metrics4}
\end{figure}

Figures~\ref{fig:metrics1}, \ref{fig:metrics2}, \ref{fig:metrics3}, and \ref{fig:metrics4} are analogues of Figure~\ref{fig:global} where the best parameters for only one of the experimental condition is given. For instance, Figure~\ref{fig:metrics1} gives the perplexity ratios and the NDCG$_5$ on the two evaluation sets for the parameters allowing the best perplexity ratio on $S_{test}$.

\subsection*{Influence of WA negative weights}
A known and intensively studied~\cite{deni08,bail11,balle14,splearn,balle18} behavior of Weighted Automata is their ability to assign negative weights to some strings. This is the counter-part of their great expressive power: when the Hankel matrix is not complete, or when its rank is not finite, or when its values are too noisy, the obtained WA might not represent exactly a probability distribution. It is important to understand that the absolute value of a negative weight does not carry any semantic: a negative weight for a WA approximating a probability distribution is exactly equivalent to having a probability of $0$.

This has no impact on the computation of NDCG, but it causes problem for the one of the perplexity since one needs to compute $\log(P_{WA}(w))$. As it cannot be replaced by $0$ either, we follow a commonly accepted path and chose to replace all negative values by a tiny $\epsilon$. In the experiments presented here, we set $\epsilon=10^{-30}$.% (this is smaller from an important factor to the smallest probability given by the RNN on a element of an evaluation set).

\begin{figure}[ht]
\begin{center}
\includegraphics[width=\textwidth]{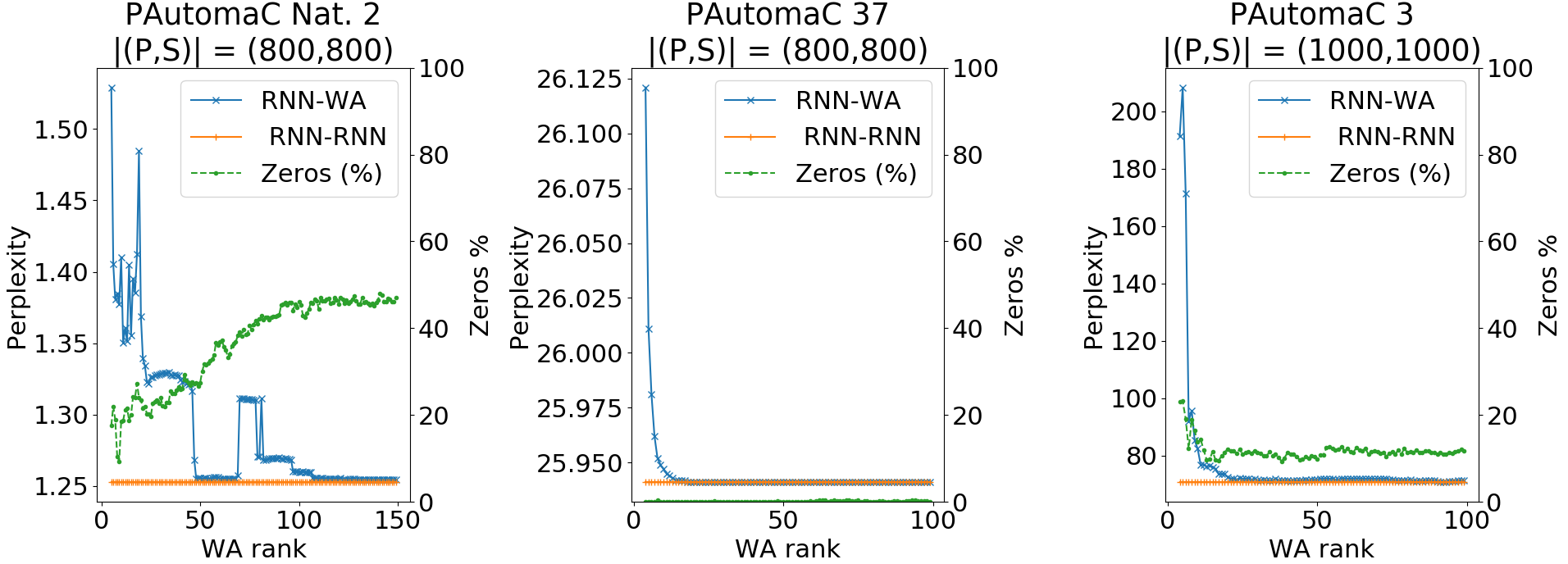}
\end{center}
\caption{Influence of the rank parameter on the perplexity of 3 datasets, together with the number of zeros.}
\label{fig:zeros}
\end{figure}

\begin{figure}[ht]
\begin{center}
\includegraphics[width=\textwidth]{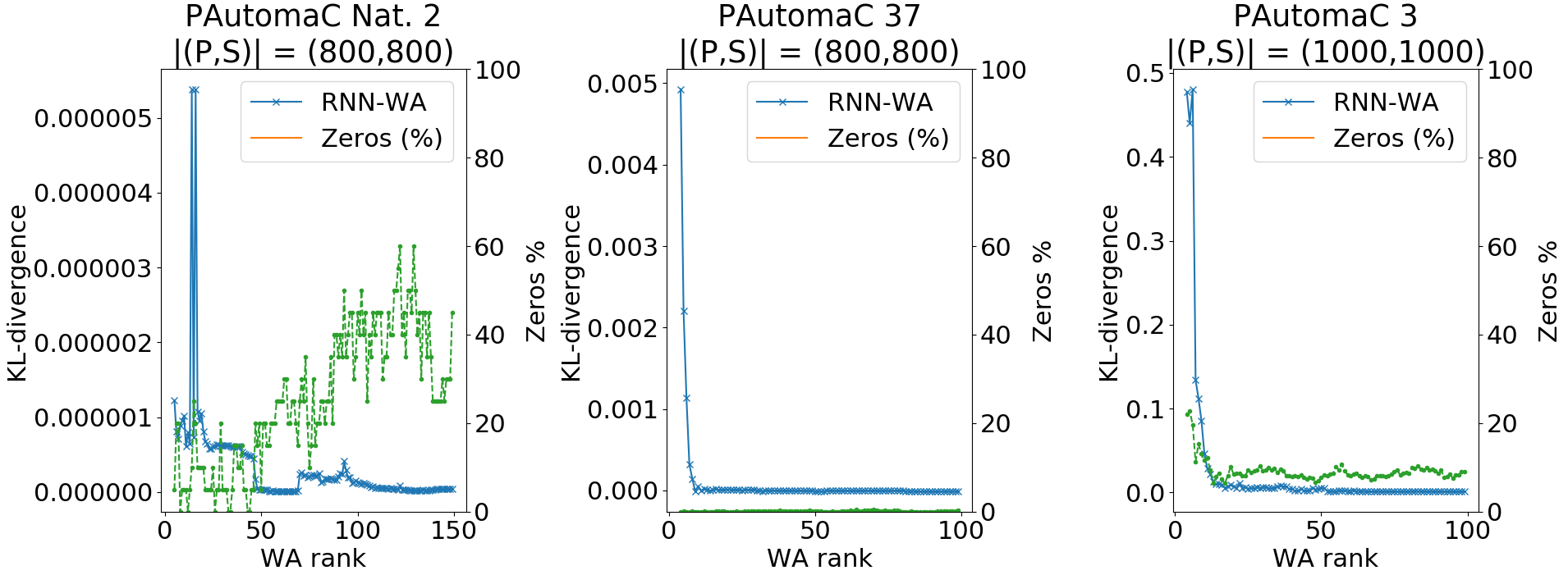}
\end{center}
\caption{Influence of the rank parameter on the Kullback-Leibler divergence of 3 datasets, together with the number of zeros.}
\label{fig:KL-zeros}
\end{figure}

Figures~\ref{fig:zeros} and \ref{fig:KL-zeros} gives the evolution of the perplexity and the KL-divergence, respectively, when the rank increases on 3 different datasets, together with the percentage of epsilon use. On PAutomaC 37, almost no element of the evaluation is given a negative weight, whatever the rank is. 
On PAutomaC 3, the number of zeros rapidly decreases and tends to stabilize around $10\%$, which is an usually accepted rate.

On PAutomaC Natural Problem 2, the number of zeros increases with the rank, finishing above the $40\%$ rate for large rank values. However, the perplexity is not affected since it stays close to the best possible perplexity (the one of the RNN, given by the flat orange line). The natural explanation is that epsilons are given to extremely unlikely strings in the RNN. Indeed, if the WA assigns a negative weight to a string $w$, $\epsilon$ is use for $P_{WA}(w)$, which means that $log(P_{WA})$ is negative with a huge absolute value. As $\mbox{Perplexity}(\{w\})=2^{-P_{RNN}(w)log(P_{WA})}$, if $P_{RNN}(w)$ is not exceptionally small, then this perplexity would be extremely high, inexorably damaging the overall perplexity (it is not uncommon to witness models with perplexity over a million on some benchmarks). The fact that it does not happen here implies that $P_{RNN}(w)$ has to be small for strings of negative weights. Therefore, by assigning a zero probability to these strings, the WA realizes a good approximation of the RNN.   

\subsection*{WER and NDCG$_1$}
\begin{figure}[ht]
\begin{center}
\includegraphics[width=\textwidth]{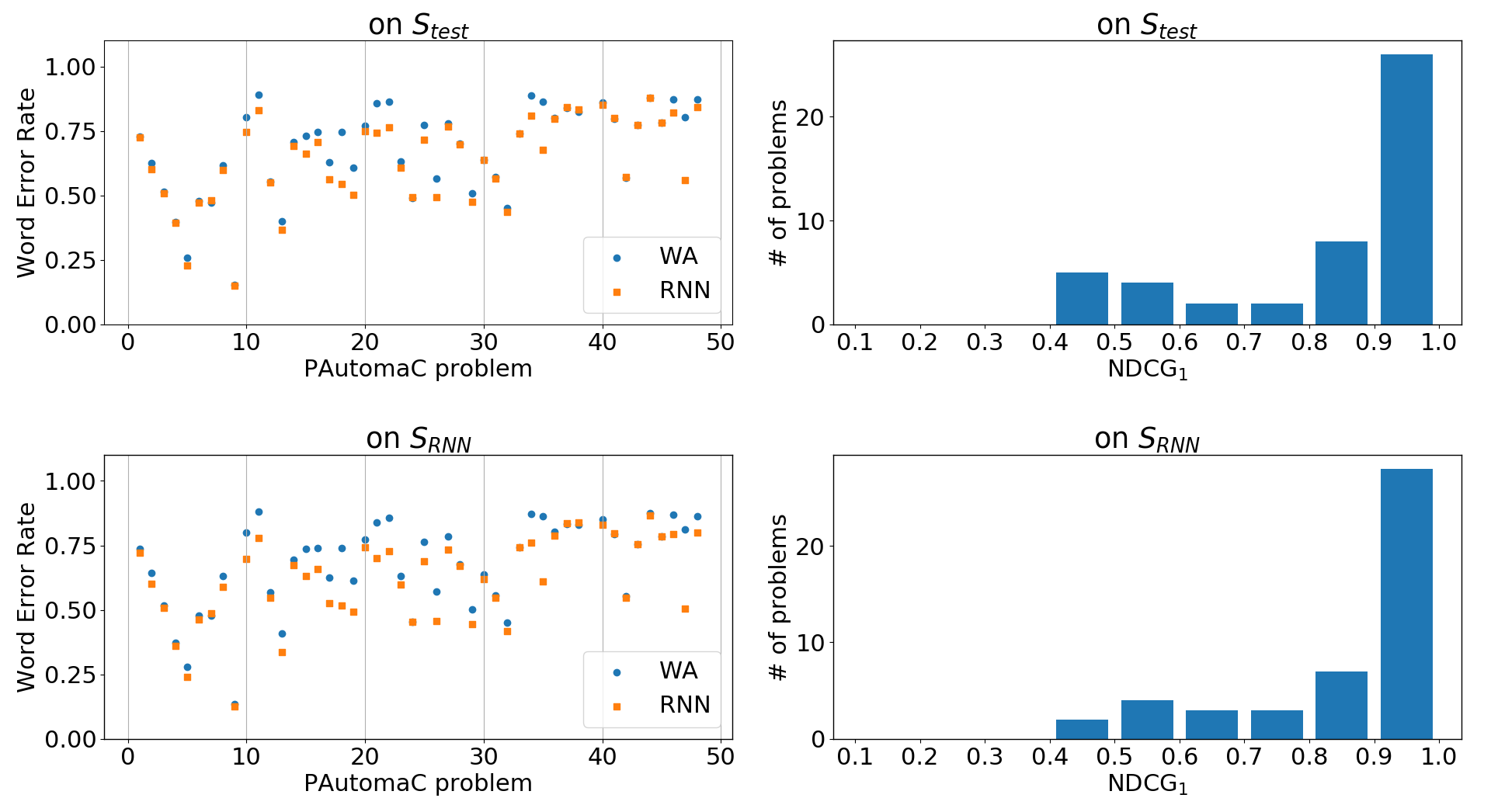}
\end{center}
\caption{Results of WER and NDCG$_1$ with best parameters for each task and each PAutomaC problem.}
\label{fig:WER}
\end{figure}

Figure~\ref{fig:WER} shows the best WER and NDCG$_1$ obtained on the evaluation sets on the synthetic problems. 

\FloatBarrier
\subsection*{Best extraction parameters on PAutomaC}
Tables~\ref{table:best-Stest} and \ref{table:best-SRNN} give, for each problem of the PAutomaC dataset, the extraction parameters that obtain the best corresponding metric on $S_{test}$ and $S_{RNN}$, respectively. Columns \textit{Rank} and $(P,S)$ contain respectively the rank value and the size of the basis that achieve the \textit{Value}. 
Column \textit{Zeros} gives the percentage of negative weights obtained when parsing a string of the evaluation set with the WA (see discussion about Figure~\ref{fig:zeros} for details on that matter).

\pagestyle{empty}
\begin{table}[ht]
\vspace{-42pt}
\centering
\begin{tabular}{|c||c|c|c|c||c|c|c|}
\hline
Pb. & \multicolumn{4}{c||}{Perplexity Ratio} & \multicolumn{3}{c|}{NDCG$_5$} \\ \hline
\# & Rank & (P,S) & Value & Zeros & Rank & (P,S) & Value \\ \hline
1 & 32 & (1400,1400) & 1.00009 & 4.2 \% & 96 & (1400,1400) & 0.95324\\ \hline
2 & 10 & (1400,1400) & 1.00162 & 2.2 \% & 12 & (1400,1400) & 0.92548\\ \hline
3 & 71 & (800,800) & 1.00012 & 5.0 \% & 75 & (1400,1400) & 0.99268\\ \hline
4 & 79 & (800,800) & 1.00073 & 2.1 \% & 42 & (800,800) & 0.99615\\ \hline
5 & 49 & (800,800) & 1.01140 & 0.6 \% & 51 & (800,800) & 0.98552\\ \hline
6 & 80 & (1400,1400) & 1.00114 & 6.5 \% & 57 & (1400,1400) & 0.98719\\ \hline
7 & 18 & (800,800) & 1.00296 & 0.0 \% & 17 & (800,800) & 0.99654\\ \hline
8 & 100 & (1000,1000) & 0.98765 & 15.2 \% & 100 & (1400,1400) & 0.91531\\ \hline
9 & 16 & (1400,1400) & 0.36845 & 24.6 \% & 19 & (1400,1400) & 0.79179\\ \hline
10 & 96 & (1000,1000) & 0.99008 & 30.5 \% & 39 & (1400,1400) & 0.71537\\ \hline
11 & 99 & (1400,1400) & 0.98284 & 36.4 \% & 98 & (1400,1400) & 0.61458\\ \hline
12 & 60 & (1000,1000) & 1.00044 & 8.6 \% & 35 & (1400,1400) & 0.97362\\ \hline
13 & 100 & (1400,1400) & 0.99040 & 8.0 \% & 58 & (1400,1400) & 0.97375\\ \hline
14 & 83 & (800,800) & 0.99967 & 5.0 \% & 47 & (1000,1000) & 0.94324\\ \hline
15 & 62 & (1000,1000) & 0.99979 & 23.0 \% & 32 & (1400,1400) & 0.76211\\ \hline
16 & 96 & (1000,1000) & 0.99283 & 19.3 \% & 88 & (1000,1000) & 0.83098\\ \hline
17 & 27 & (1000,1000) & 0.99520 & 14.3 \% & 28 & (1000,1000) & 0.83255\\ \hline
18 & 88 & (1400,1400) & 0.74500 & 24.2 \% & 95 & (1000,1000) & 0.59521\\ \hline
19 & 96 & (1400,1400) & 0.99375 & 18.3 \% & 100 & (1400,1400) & 0.83020\\ \hline
20 & 16 & (1000,1000) & 1.00102 & 30.1 \% & 65 & (1000,1000) & 0.90750\\ \hline
21 & 12 & (800,800) & 1.04705 & 37.1 \% & 69 & (1400,1400) & 0.54177\\ \hline
22 & 77 & (1400,1400) & 0.97386 & 36.6 \% & 97 & (800,800) & 0.57218\\ \hline
23 & 56 & (1400,1400) & 1.00000 & 9.4 \% & 76 & (1400,1400) & 0.94410\\ \hline
24 & 15 & (800,800) & 1.00030 & 0.0 \% & 39 & (800,800) & 0.99999\\ \hline
25 & 99 & (800,800) & 0.99996 & 17.0 \% & 95 & (1400,1400) & 0.81833\\ \hline
26 & 100 & (1000,1000) & 0.97447 & 14.6 \% & 89 & (1400,1400) & 0.88117\\ \hline
27 & 27 & (1400,1400) & 1.00174 & 11.9 \% & 21 & (1000,1000) & 0.87799\\ \hline
28 & 65 & (1400,1400) & 1.00000 & 3.3 \% & 95 & (1400,1400) & 0.98394\\ \hline
29 & 71 & (1400,1400) & 0.99554 & 7.8 \% & 95 & (1400,1400) & 0.95355\\ \hline
30 & 21 & (1000,1000) & 1.00051 & 3.5 \% & 43 & (1400,1400) & 0.96598\\ \hline
31 & 38 & (800,800) & 1.00000 & 3.2 \% & 68 & (1400,1400) & 0.98993\\ \hline
32 & 86 & (1000,1000) & 0.99836 & 5.1 \% & 98 & (1400,1400) & 0.97969\\ \hline
33 & 8 & (1400,1400) & 1.00046 & 0.9 \% & 71 & (1000,1000) & 0.99381\\ \hline
34 & 96 & (1400,1400) & 0.99867 & 41.0 \% & 97 & (1400,1400) & 0.57356\\ \hline
35 & 84 & (800,800) & 0.93308 & 37.3 \% & 77 & (1400,1400) & 0.44417\\ \hline
36 & 14 & (1000,1000) & 1.00008 & 10.0 \% & 91 & (1400,1400) & 0.95274\\ \hline
37 & 11 & (800,800) & 1.00000 & 0.0 \% & 98 & (1400,1400) & 0.99914\\ \hline
38 & 5 & (1000,1000) & 1.00028 & 1.5 \% & 82 & (1400,1400) & 0.99774\\ \hline
40 & 84 & (1000,1000) & 0.99959 & 24.0 \% & 99 & (1400,1400) & 0.82376\\ \hline
41 & 6 & (800,800) & 1.00007 & 0.0 \% & 85 & (800,800) & 0.99998\\ \hline
42 & 8 & (1000,1000) & 1.00158 & 0.2 \% & 19 & (800,800) & 0.99515\\ \hline
43 & 11 & (800,800) & 1.00003 & 0.0 \% & 74 & (800,800) & 0.99999\\ \hline
44 & 5 & (1000,1000) & 1.00064 & 2.4 \% & 39 & (1000,1000) & 0.94511\\ \hline
45 & 2 & (800,800) & 1.00021 & 0.4 \% & 62 & (1400,1400) & 0.99904\\ \hline
46 & 33 & (1400,1400) & 1.00143 & 34.2 \% & 64 & (1400,1400) & 0.64332\\ \hline
47 & 84 & (1400,1400) & 0.93862 & 31.3 \% & 57 & (1400,1400) & 0.48529\\ \hline
48 & 15 & (800,800) & 1.00154 & 23.5 \% & 52 & (1400,1400) & 0.72140\\ \hline
\end{tabular}
\caption{Parameters for best obtained WA for each PAutomaC problems on $S_{test}$}
\label{table:best-Stest}
\end{table}

\begin{table}[ht]
\vspace{-42pt}
\centering
\begin{tabular}{|c||c|c|c|c||c|c|c|}
\hline
Pb. & \multicolumn{4}{c||}{Perplexity Ratio} & \multicolumn{3}{c|}{NDCG$_5$} \\ \hline
\# & Rank & (P,S) & Value & Zeros & Rank & (P,S) & Value \\ \hline
1 & 90 & (1400,1400) & 0.99995 & 6.8 \% & 91 & (1400,1400) & 0.95030\\ \hline
2 & 65 & (800,800) & 0.99972 & 3.1 \% & 12 & (1400,1400) & 0.92548\\ \hline
3 & 92 & (1000,1000) & 0.99756 & 7.7 \% & 98 & (1400,1400) & 0.99232\\ \hline
4 & 94 & (800,800) & 0.99901 & 1.9 \% & 54 & (1400,1400) & 0.99407\\ \hline
5 & 32 & (1000,1000) & 0.99064 & 0.1 \% & 51 & (800,800) & 0.98552\\ \hline
6 & 100 & (1400,1400) & 0.99912 & 3.1 \% & 35 & (1400,1400) & 0.98669\\ \hline
7 & 29 & (800,800) & 0.98952 & 0.6 \% & 17 & (800,800) & 0.99654\\ \hline
8 & 100 & (1400,1400) & 0.97525 & 12.5 \% & 98 & (1400,1400) & 0.91287\\ \hline
9 & 16 & (1400,1400) & 0.31197 & 24.6 \% & 22 & (1400,1400) & 0.79165\\ \hline
10 & 70 & (1400,1400) & 0.97911 & 32.0 \% & 40 & (1400,1400) & 0.71336\\ \hline
11 & 66 & (1400,1400) & 0.98303 & 34.4 \% & 88 & (1400,1400) & 0.60401\\ \hline
12 & 93 & (1400,1400) & 0.99879 & 7.2 \% & 34 & (1400,1400) & 0.96598\\ \hline
13 & 89 & (1000,1000) & 0.96508 & 10.4 \% & 56 & (1400,1400) & 0.96758\\ \hline
14 & 53 & (1400,1400) & 0.99683 & 7.0 \% & 9 & (1000,1000) & 0.93522\\ \hline
15 & 100 & (1400,1400) & 0.99204 & 25.6 \% & 28 & (1400,1400) & 0.75812\\ \hline
16 & 97 & (1400,1400) & 0.96902 & 25.0 \% & 97 & (1000,1000) & 0.82633\\ \hline
17 & 72 & (1000,1000) & 0.97144 & 25.6 \% & 26 & (1400,1400) & 0.80701\\ \hline
18 & 88 & (1400,1400) & 0.63316 & 24.2 \% & 95 & (1000,1000) & 0.59521\\ \hline
19 & 99 & (800,800) & 0.97982 & 23.2 \% & 100 & (1400,1400) & 0.83020\\ \hline
20 & 74 & (800,800) & 0.98953 & 30.8 \% & 98 & (1000,1000) & 0.88840\\ \hline
21 & 85 & (1000,1000) & 0.96581 & 39.9 \% & 69 & (1400,1400) & 0.54177\\ \hline
22 & 95 & (1000,1000) & 0.96631 & 39.7 \% & 100 & (1000,1000) & 0.56973\\ \hline
23 & 57 & (800,800) & 0.99905 & 13.3 \% & 68 & (1400,1400) & 0.93060\\ \hline
24 & 24 & (1000,1000) & 0.99993 & 0.0 \% & 32 & (800,800) & 0.99997\\ \hline
25 & 95 & (1000,1000) & 0.99921 & 19.6 \% & 99 & (1400,1400) & 0.81832\\ \hline
26 & 90 & (1000,1000) & 0.96623 & 14.1 \% & 81 & (1400,1400) & 0.87139\\ \hline
27 & 100 & (1000,1000) & 0.99796 & 18.6 \% & 63 & (1400,1400) & 0.85724\\ \hline
28 & 87 & (800,800) & 0.99932 & 12.0 \% & 97 & (1400,1400) & 0.98082\\ \hline
29 & 81 & (1400,1400) & 0.97509 & 7.3 \% & 97 & (1400,1400) & 0.94998\\ \hline
30 & 91 & (1400,1400) & 0.99976 & 5.3 \% & 80 & (1000,1000) & 0.95873\\ \hline
31 & 26 & (1000,1000) & 0.99650 & 4.3 \% & 95 & (1400,1400) & 0.98882\\ \hline
32 & 97 & (1000,1000) & 0.99303 & 9.5 \% & 99 & (1400,1400) & 0.97968\\ \hline
33 & 73 & (1000,1000) & 0.99998 & 11.5 \% & 77 & (1000,1000) & 0.98977\\ \hline
34 & 100 & (1400,1400) & 0.97584 & 37.4 \% & 92 & (1400,1400) & 0.56693\\ \hline
35 & 87 & (1400,1400) & 0.82007 & 40.2 \% & 62 & (800,800) & 0.44097\\ \hline
36 & 57 & (800,800) & 0.99997 & 11.3 \% & 100 & (1400,1400) & 0.95238\\ \hline
37 & 18 & (800,800) & 1.00000 & 0.1 \% & 100 & (1400,1400) & 0.99914\\ \hline
38 & 19 & (1400,1400) & 1.00000 & 7.2 \% & 93 & (1400,1400) & 0.99760\\ \hline
40 & 75 & (800,800) & 0.99560 & 27.6 \% & 100 & (1400,1400) & 0.82201\\ \hline
41 & 15 & (800,800) & 1.00000 & 0.0 \% & 83 & (800,800) & 0.99997\\ \hline
42 & 80 & (800,800) & 0.99968 & 6.4 \% & 7 & (1000,1000) & 0.99442\\ \hline
43 & 17 & (1400,1400) & 1.00000 & 0.0 \% & 72 & (800,800) & 0.99998\\ \hline
44 & 24 & (800,800) & 1.00000 & 6.1 \% & 99 & (1000,1000) & 0.94282\\ \hline
45 & 19 & (1400,1400) & 1.00000 & 0.2 \% & 4 & (1400,1400) & 0.99885\\ \hline
46 & 46 & (1000,1000) & 0.99110 & 37.2 \% & 72 & (1400,1400) & 0.64257\\ \hline
47 & 100 & (1400,1400) & 0.79781 & 33.7 \% & 57 & (1400,1400) & 0.48529\\ \hline
48 & 51 & (800,800) & 0.99561 & 37.6 \% & 50 & (1400,1400) & 0.71464\\ \hline
\end{tabular}
\caption{Parameters for best obtained WA for each PAutomaC problems on $S_{RNN}$}
\label{table:best-SRNN}
\end{table}

\end{document}